\documentclass[10pt,twocolumn,letterpaper]{article}

\usepackage[pagenumbers]{cvpr}
\definecolor{cvprblue}{rgb}{0.21,0.49,0.74}
\usepackage[pagebackref,breaklinks,colorlinks,allcolors=cvprblue]{hyperref}

\def\paperID{} 
\def\confName{CVPR}
\def\confYear{2026}

\title{Model-Centric Diagnostics: A Framework for Internal State Readouts}

\author{Fangzheng Wu\\
Tulane University\\
New Orleans, LA\\
{\tt\small fwu6@tulane.edu}
\and
Brian Summa\\
Tulane University\\
New Orleans, LA\\
{\tt\small bsumma@tulane.edu}
}
\PassOptionsToPackage{nameinlink,capitalize}{cleveref}

\usepackage{mdframed}
\usepackage{amsmath,amssymb}
\usepackage{graphicx}
\usepackage{booktabs}
\usepackage{multirow}
\usepackage{siunitx}

\usepackage[table]{xcolor}

\usepackage{textcomp}
\usepackage{enumitem}

\usepackage{threeparttable}

\usepackage{caption}
\usepackage{subcaption}

\usepackage{cleveref}

\usepackage{indentfirst}  

\crefname{section}{Section}{Sections}
\Crefname{section}{Section}{Sections}
\crefname{figure}{Fig.}{Figs.}
\Crefname{figure}{Figure}{Figures}
\crefname{table}{Table}{Tables}
\Crefname{table}{Table}{Tables}

\usepackage{siunitx,booktabs,threeparttable}
\newcommand{\ggrad}{\|g\|_F}

\makeatletter
\renewcommand\paragraph{\@startsection{paragraph}{4}{\z@}%
  {0.5ex \@plus .2ex}
  {-0.8em}
  {\normalfont\normalsize\itshape}}
\makeatother

\begin{document}
\maketitle

\renewcommand{\thefootnote}{\fnsymbol{footnote}}
\footnotetext[1]{\textbf{Disclaimer:} This manuscript presents a conceptual framework and differs substantially from a forthcoming submission that provides full algorithmic details, theoretical analysis, and comprehensive experimental validation. The preliminary results shown here are illustrative only.}
\renewcommand{\thefootnote}{\arabic{footnote}}
\setcounter{footnote}{0}

\begin{abstract}
We present a model-centric diagnostic framework that treats training state as a latent variable and unifies a family of internal readouts---head-gradient norms, confidence, entropy, margin, and related signals---as anchor-relative projections of that state.
A preliminary version of this work introduced a head-gradient probe for checkpoint selection.
In this version, we focus on the unifying perspective and structural diagnostics; full algorithmic details, theoretical analysis, and experimental validation will appear in a forthcoming paper.

We outline the conceptual scaffold: any prediction head induces a local loss landscape whose geometry (gradient magnitude, curvature, sharpness) reflects how well the upstream features are aligned with the task.
Different readout choices---gradient norms, softmax entropy, predictive margin---correspond to different projections of this geometry, each with complementary strengths.
The framework suggests that checkpoint selection, early stopping, and lightweight architecture pre-screening can all be viewed as querying the same underlying state through different lenses.

Illustrative experiments on ImageNet classification and COCO detection/segmentation hint at the practical potential; rigorous benchmarks and ablations are deferred to the full paper.

\end{abstract}

\section{Introduction}

Modern training loops heavily lean on held-out validation sets to decide when to stop and which checkpoint or model to keep. This practice costs extra compute for periodic evaluation, can be impossible when labels are scarce or private, and may be unreliable under distribution shift. We revisit a minimalistic question: \emph{can one predict model quality using internal state readouts---signals derived from the model itself---without any validation data at selection time?}

\paragraph{A unifying perspective.}
We argue that seemingly disparate diagnostics---gradient norms, softmax confidence, entropy, margin, and curvature-based measures---can all be understood as \emph{anchor-relative projections} of a latent training state.
When features become more linearly separable, the classifier head operates in a flatter local landscape.
This geometry manifests through multiple observable proxies: smaller head gradients, higher confidence, lower entropy, and wider margins.
Each readout offers a different ``lens'' into the same underlying state.

\paragraph{Framework contributions.}
This paper sketches a \textbf{model-centric diagnostic framework} with the following elements:
\begin{itemize}[leftmargin=1.2em]
  \item \textbf{Training state as latent variable.} We treat the quality of learned representations as a hidden variable that causally influences multiple observable readouts.
  \item \textbf{Anchor-relative projections.} Each diagnostic (gradient norm, confidence, etc.) projects the latent state onto a scalar or low-dimensional summary, analogous to different projections of a high-dimensional object.
  \item \textbf{Task-agnostic structure.} The same conceptual scaffold applies to classification, detection, segmentation, and generative modeling, though the optimal readout may differ by task.
\end{itemize}

\paragraph{Scope of this version.}
A preliminary version of this work focused on one specific instantiation: the head-gradient norm as a checkpoint selection signal.
Here, we step back to emphasize the \emph{unifying framework} rather than any single method.
We include illustrative experiments to motivate the perspective, but defer rigorous benchmarks, ablations, and theoretical analysis to a forthcoming full paper.

\paragraph{Why this matters.}
If internal readouts reliably track model quality, practitioners gain several benefits: validation-free early stopping, lightweight pre-screening of architectures, and monitoring in settings where labels are unavailable or expensive.
The framework perspective suggests that these applications share a common foundation and that progress on one readout informs the others.

\section{Related Work}

A large line of work aims to rank architectures or checkpoints
without (or before) full training using ``zero-cost'' or
training-free indicators computed from a few forward/backward
passes at (near) initialization. Representative examples include
linear-region/Zen-based measures~\citep{zenscore},
trainability indicators (e.g., TE-NAS~\cite{tenas}, NWOT~\cite{nwot}, and gradient-flow proxies such as SynFlow~\citep{synflow}.
These methods are chiefly designed for \emph{architecture search}:
they compare many untrained candidates in a fixed search space and
correlate the proxy with the eventual trained accuracy.
Our setting is complementary: we operate on \emph{running checkpoints
or off-the-shelf pre-trained models}, and the goal is \emph{early
stopping / model or checkpoint selection} rather than universal
ranking inside a NAS space.

Several NAS papers~\citep{abdelfattah2021zerocostproxieslightweightnas,colin2022adeeperlook} report a baseline termed \emph{grad-norm} that
measures the $\ell_p$-norm of the \emph{full-parameter} gradient
$\|\nabla_{\theta}\mathcal{L}\|$ near initialization to score
architectures without training (see the comparisons in
ZenNAS~\citep{zenscore}).
Our probe is different in both \emph{where} and \emph{what} it
measures: (i) we evaluate at \emph{trained} or \emph{training} checkpoints
rather than at random initialization; (ii) we backpropagate through the
\emph{classifier head only} and \emph{detach features}, producing a signal
that directly reflects the linear separability of current features.
Empirically, this one-batch, head-only signal is competitive while being
simpler and cheaper (no search-space–specific derivations).

Connections between loss landscape geometry and generalization have
been studied via flatness/sharpness and Hessian spectra
(e.g.,~\citealp{keskar2016large,jiang2019fantastic,hochreiter1997flatminima,foret2021sam}).
Large gradients often co-occur with sharper regions of the loss;
our findings align with this picture: models/checkpoints with smaller
\emph{head} gradients tend to exhibit better Top-1 and lower loss,
suggesting improved separability of the learned features under the
current head.

Prior attempts to bypass a held-out validation set include curve
extrapolation of the training trajectory, self-supervised or
auxiliary criteria, and proxy losses
(e.g.,~\citealp{domhan2015learningcurve,klein2017lcp,swersky2014freezethaw,mahsereci2017earlystopping,achille2019task2vec}).
Compared to those, our approach is \emph{instant} (one forward+backward
on a single supervised mini-batch), requires \emph{no extra data}, and
is architecture-agnostic. We also provide ablations against alternative
head-only signals (L1/L2/L$\infty$, Fisher trace~\cite{amari1998natural}, and scale-normalized
variants), and show that a plain $\ell_2$ head-gradient norm offers a
strong accuracy proxy with minimal complexity.

While our focus is validation-free checkpoint/model selection,
the same probe can serve as a lightweight pre-screening module in NAS:
quickly filter a pool of trained-once or partially trained candidates
before expensive fine-tuning. This use is orthogonal to training-free
NAS proxies~\citep{zenscore,tenas,nwot,synflow} and avoids assuming a search space.

\section{Framework: Internal State Readouts}
\label{sec:framework}

This section describes the conceptual framework that unifies various internal diagnostics as projections of training state.
We present one concrete instantiation---the head-gradient norm---as an illustrative example; a complete treatment with theoretical analysis and systematic comparisons will appear in the full paper.

\subsection{Training State as Latent Variable}

We posit that the ``quality'' of learned representations at any checkpoint can be modeled as a latent variable $\mathcal{S}$ that causally influences multiple observables.
As training progresses and features become more linearly separable, this latent state improves, and various readouts reflect that improvement:
\begin{itemize}[leftmargin=1.2em]
  \item \textbf{Gradient-based:} Head-gradient norms, full-model gradient norms, Fisher information traces
  \item \textbf{Output-based:} Softmax confidence, predictive entropy, margin between top-2 classes
  \item \textbf{Geometry-based:} Loss curvature, sharpness measures, Hessian eigenvalues
\end{itemize}
Each readout is a different \emph{projection} of $\mathcal{S}$---analogous to viewing a 3D object from different angles.

\subsection{Why Gradient Norms?}

The gradient norm at a checkpoint reflects the local loss curvature~\cite{keskar2016large,jiang2019fantastic}.
For a model with parameters $\theta$ at checkpoint $t$:
\begin{equation}
\|\nabla_\theta \mathcal{L}(\theta_t)\| \approx \lambda_{\max}(H_t) \cdot \|\theta_t - \theta^*\|
\end{equation}
where $H_t$ is the Hessian at $\theta_t$, $\lambda_{\max}$ is its maximum eigenvalue (sharpness), and $\theta^*$ is a local minimum.

The connection to generalization is well-established: flat minima (low sharpness) generalize better than sharp minima.
The head-gradient norm, in particular, measures how ``settled'' the final classification layer is given the upstream features---a direct proxy for feature quality.

\subsection{Example Instantiation: Head-Gradient Probe}
\label{sec:head-grad-example}

As one concrete instance of the framework, we describe a head-gradient probe that has shown promise in preliminary experiments.
\emph{This is presented as an illustrative example; the full paper will provide systematic comparisons across multiple readout types.}

\paragraph{Notation.}
Let $f_\theta(x)=h_W(\phi_\psi(x))$, where $\phi_\psi:\mathcal{X}\!\to\!\mathbb{R}^d$
is a feature extractor and $h_W(z)=Wz$ is a linear head with $W\in\mathbb{R}^{C\times d}$.
Given a mini-batch $\{(x_i,y_i)\}_{i=1}^B$, define the \emph{detached} feature matrix
$Z=[\phi_\psi(x_1),\ldots,\phi_\psi(x_B)]\in\mathbb{R}^{d\times B}$,
the probabilities $P=\mathrm{softmax}(WZ)$,
and the cross-entropy $\mathcal{L}=-\tfrac{1}{B}\sum_{i=1}^{B}\log P_{y_i,i}$.
The head-only gradient norm is:
\begin{equation}
\label{eq:headgrad}
\|g\|_F \triangleq \big\|\nabla_W \mathcal L\big\|_{F} = \Big\|\frac{1}{B}\,(P-Y)\,Z^\top\Big\|_F.
\end{equation}
Features are \emph{detached} so that gradients do not flow into the backbone $\phi_\psi$.

\paragraph{Scale normalization variants.}
To improve comparability across architectures, one can normalize by feature or head magnitude:
\begin{equation}
\mathrm{score}_z = \frac{\|g\|_F}{\|Z\|_F + \varepsilon}, \quad
\mathrm{score}_w = \frac{\|g\|_F}{\|W\|_F + \varepsilon_w}.
\end{equation}
Preliminary results suggest feature-scale normalization works well for Transformers/modern CNNs, while head-scale normalization is often more stable within classic CNN families.

\paragraph{Selection rule.}
Given checkpoints $\{\theta_t\}$ from one run, a simple validation-free rule is:
$\hat{t}=\arg\min_t \|g\|_F(\theta_t)$.
The selection itself never uses validation labels.

\paragraph{Cost.}
The probe backpropagates only through the head: $\mathcal{O}(CdB)$ FLOPs for $W\in\mathbb{R}^{C\times d}$ and batch size $B$---negligible relative to a full validation pass.

\section{Experimental Setup}
\label{sec:setup}

We evaluate our method mainly on 4 tasks: image classification, object detection, instance segmentaion and standard UNet/DDPM.

\subsection{Image Classification on ImageNet}

\textbf{Architectures.}
We evaluate \textbf{25} ImageNet-1k classifiers spanning modern CNNs and Transformers.
To keep the main narrative focused, figures highlight seven representative backbones:
ResNet50~\cite{he2016deep}, EfficientNet-B0~\cite{tan2019efficientnet},
ViT-Base/Small~\cite{dosovitskiy2020image}, DeiT-Small~\cite{touvron2021training},
Swin-Tiny~\cite{liu2021swin}, and MobileNetV3-Large~\cite{howard2019mobilenetv3}.
ConvNeXt-Tiny~\cite{liu2022convnet} appears in selected ablations.
The  roster of 25 models with family breakdown and parameter counts is in supplementary material.

\textbf{Dataset.} ImageNet-1k (ILSVRC2012)~\cite{russakovsky2015imagenet} with standard preprocessing: random resized crop (224$\times$224), random horizontal flip, and normalization.

\textbf{Training.} All models are trained for 200 epochs using standard recipes. Checkpoints are saved every 5 steps for gradient measurement.

\textbf{Evaluation Metrics.} We compute Pearson correlation $r$ between the head-gradient norm $\|\nabla_W\mathcal L\|_F$ and validation Top-1 (or loss) over the training trajectory.
 The default window is $n{=}100$ steps. For robustness, we vary the EMA (exponential moving average)~\cite{brown1959exp} parameter $k \in \{1,3,5,9\}$ and last-steps window size.

\textbf{Baselines.} We compare against confidence-based and margin-based~\cite{guo2017calibration} checkpoint selection methods.

\subsection{Object Detection on COCO}

We use a light-weight assignment ($IoU>0.5$ nearest) to isolate classification-head
gradients; this avoids backpropagating through full Hungarian matching but
introduces assignment noise. 

\textbf{Detectors.} We evaluate five diverse detectors from TorchVision: Faster R-CNN~\cite{ren2015faster} (ResNet50 and MobileNetV3 backbones), RetinaNet~\cite{lin2017focal} (ResNet50), SSD300~\cite{liu2016ssd} (VGG16), and SSDLite~\cite{sandler2018mobilenetv2} (MobileNetV3), spanning both two-stage and single-stage paradigms.

\textbf{Dataset.} COCO 2017~\cite{lin2014microsoft} with 118k training and 5k validation images.

\textbf{Gradient Computation.} For each detector, we compute the gradient norm of its classification head using 30 randomly sampled training batches (batch size 4).

\textbf{Evaluation.} Detection performance is measured as mAP (AP@IoU=0.5:0.95) on 500 validation images. We compute Pearson correlation between classification head gradient norm and mAP.

\subsection{Instance Segmentation on COCO}
\textbf{Models.} We use 12 models from three families: Mask2Former \cite{cheng2022mask2former,liu2021swin} (5 variants), TV Mask R-CNN \cite{he2017maskrcnn} (2 variants), and YOLO-Seg (YOLOv8) \cite{yaseen2024yolov8} (5 variants).

\textbf{Protocol.} Unless otherwise stated, we randomly sample 500 images from COCO val2017 to (i) compute the head-gradient $\ell_2$ norm via a single head-only backward pass on the classification head and (ii) evaluate mask AP with official COCO metrics.
We report Pearson's $r$ and Spearman's $\rho$ with nonparametric bootstrap 95\% CIs (10k resamples).

\textbf{Sensitivity.} We vary the number of evaluation images (200/500/1000) and the number of gradient micro-batches (8/16/32).

\subsection{Diffusion setup (CIFAR-10)}
\label{sec:diff-setup}
\textbf{Model \& data.}
We use a UNet/DDPM framework on CIFAR-10 ($32{\times}32$, $T{=}1000$) with a \emph{linear} $\beta$-schedule by default (cosine in ablations)~\cite{ho2020denoising,dhariwal2021diffusion,krizhevsky2009learning}.
Training uses AdamW ($10^{-4}$) for 20k steps (5k for ablations) and maintains EMA weights (decay $0.999$), which are used for sampling unless noted.

\textbf{Evaluation.}
Every 2000 steps we run the head probe on a fixed set of $B{=}256$ validation images, drawing $t\!\sim\!\mathrm{Uniform}([0.1T,0.7T])$, repeating $K{=}3$ and aggregating by median to obtain $\mathrm{score}_w$. We also log the same-distribution probe MSE.

\textbf{Tail-window selection.}
Within the last $20\%$ steps, we smooth $\mathrm{score}_w$ by EMA ($\beta{=}0.9$), take the $q{=}0.1$ quantile as candidates, apply a patience of 3 to break ties, and align by the best lead–lag within $\pm10$ steps. We report the \emph{gap to the global best} metric and compare against \texttt{Last} and \texttt{Loss-min}.

\textbf{FID protocol.}
Unless otherwise noted, we compute FID against CIFAR-10 \emph{test} (10k reals) using 2048-d Inception-v3 \texttt{avgpool} features at $299{\times}299$ with ImageNet normalization. We generate 10k (dev) or 50k (final) images with a unified sampler: DDIM ($\eta{=}0$, $N_\text{FE}{\in}\{100,250\}$) or DDPM-1000 for reference; generated images are rescaled to $[0,1]$ before feature extraction. Lower FID is better; thus higher $\mathrm{score}_w$ corresponds to lower FID (negative correlation) and better quality.

\subsection{Implementation Details}

All experiments are conducted with \textbf{PyTorch~2.9.0+cu128} and 4 NVIDIA RTX A6000 gpus. 
ImageNet experiments follow the standard training scripts from \texttt{timm}, 
while COCO experiments employ TorchVision's pretrained detection models without additional fine-tuning. 
Unless otherwise specified, a batch size of 64 is used for all ImageNet probes and accuracy evaluations. 

Unless otherwise specified we use PyTorch's \texttt{reduction='mean'}, so per-batch gradients are averaged over $B$ and the head-gradient scale is largely insensitive to batch size. Throughout, the probe backpropagates only w.r.t.\ $W$ (the classifier head), consistent with Sec.~\ref{sec:head-grad-example}.

\section{Illustrative Experiments}
\label{sec:results}

\emph{The following experiments are illustrative and motivate the framework perspective. Rigorous benchmarks with full ablations and statistical controls will appear in the forthcoming complete paper.}

\subsection{Image Classification}
\begin{figure}[t]
  \centering
  \includegraphics[width=\linewidth]{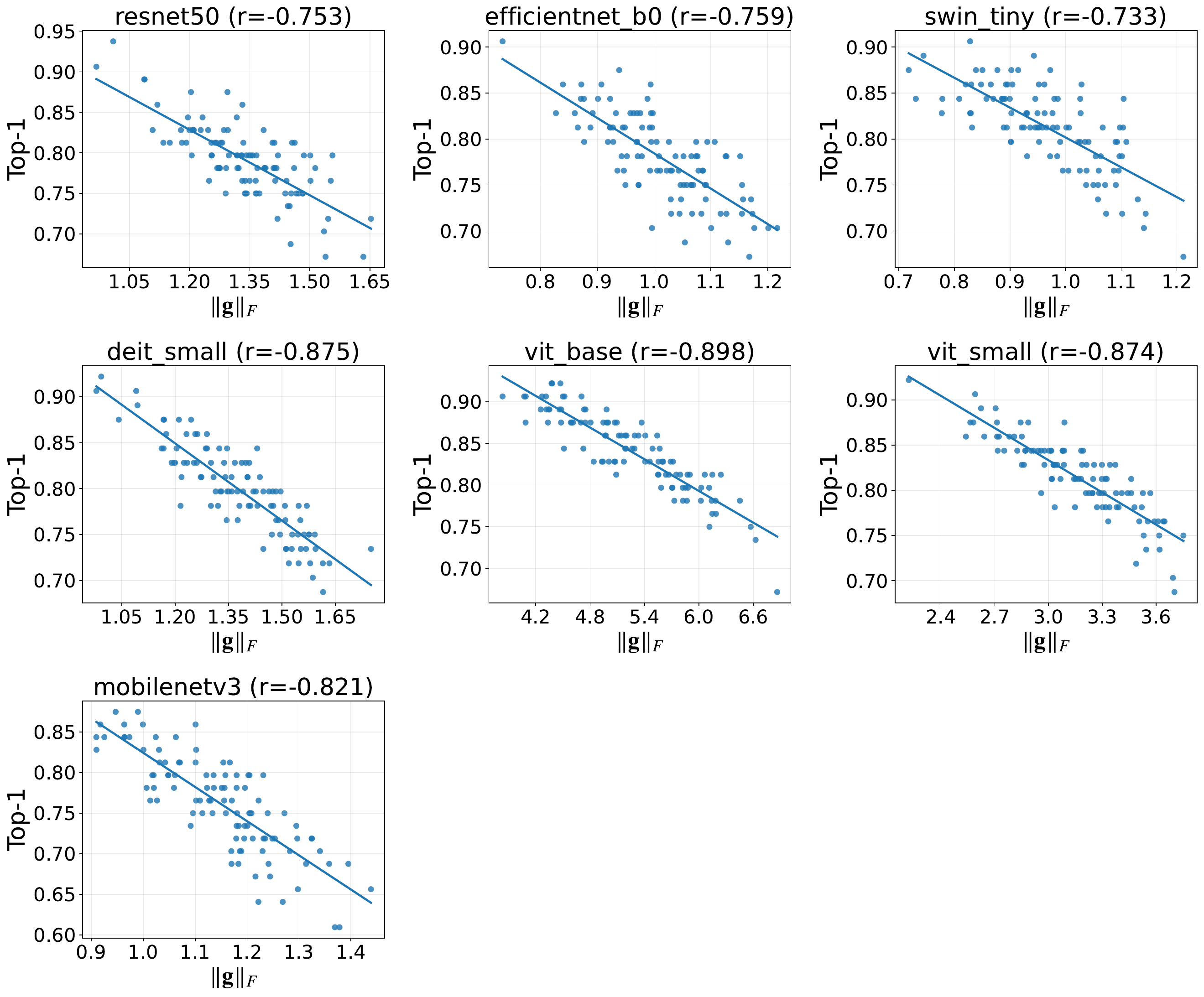}
  \caption{\textbf{Top-1 vs head-gradient norm ($\ggrad$)} on seven architectures. Lower $\ggrad$ indicates higher separability and correlates strongly (negatively) with Top-1.}
  \label{fig:scatter}
\end{figure}

Fig.~\ref{fig:scatter} shows consistent negative trends. Over seven models, we obtain $\bar r(\text{Top-1},\ggrad)=-0.854\pm0.038$ and $\bar r(\text{Loss},\ggrad)=0.884\pm0.017$. All 7 models have $|r(\text{Top-1},\ggrad)|>0.7$.
More comprehensively, across \textbf{25} ImageNet-1K models, the classifier-head gradient norm strongly anti-correlates with accuracy (Pearson $r=-0.845\pm0.052$; Spearman $r=-0.832\pm0.062$). Family-wise results remain strong: \textbf{CNNs} $r=-0.825\pm0.040$, \textbf{Transformers} $r=-0.876\pm0.033$ (Pearson). We report complete results in supplementary material.
Unless otherwise noted, we report Pearson/Spearman correlations with 95\% 
bootstrap CIs~\cite{Efron1992} (10k resamples, over models). We also ran leave-one-out (LOO)
sensitivity. The overall trend is stable and the largest LOO shift is 
$\Delta r \le 0.11$.

\begin{figure}[t]
  \centering
  \includegraphics[width=\linewidth]{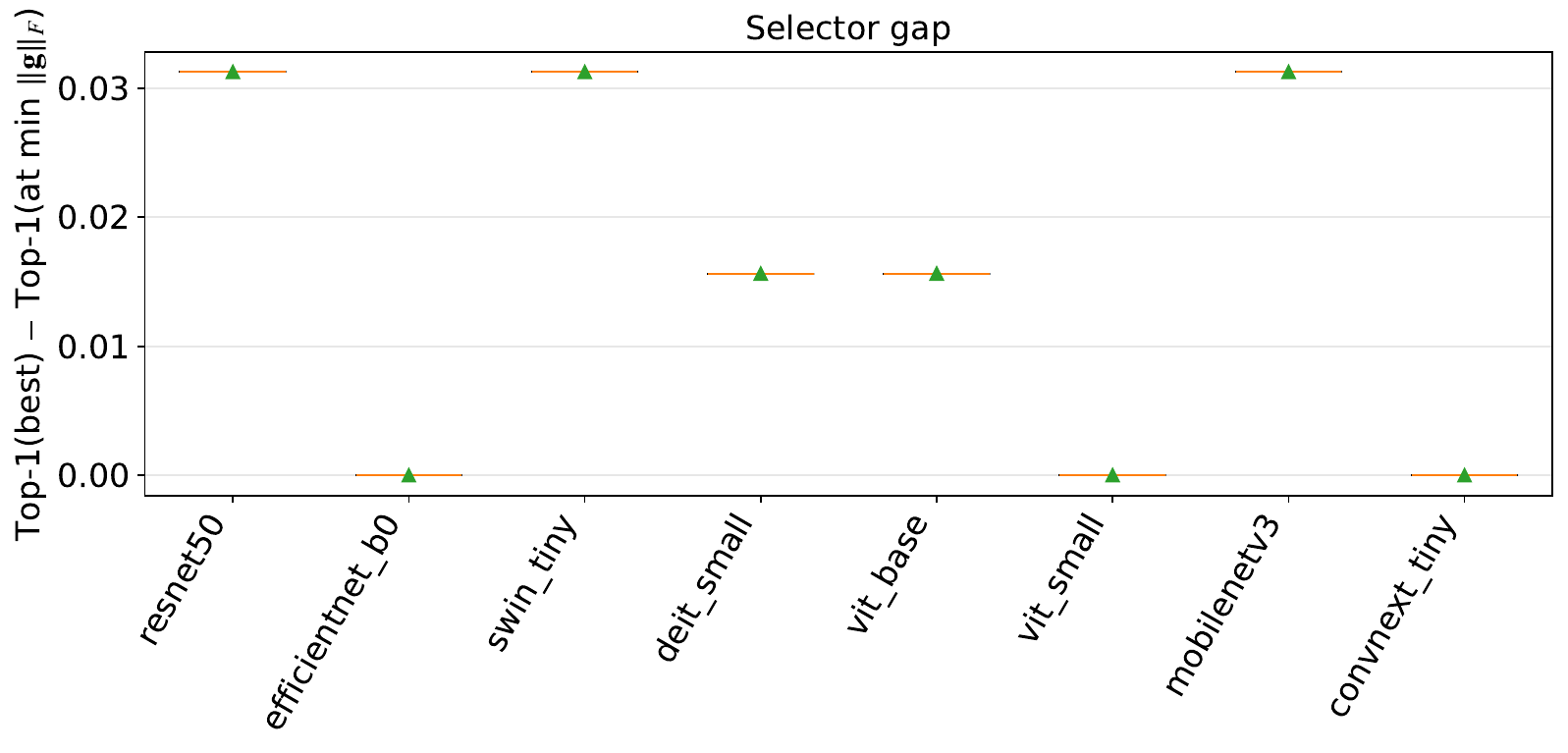}
  \caption{\textbf{Selector gap.} Top-1(best) minus Top-1(at step of minimal $\ggrad$). Most gaps are $0$–$0.03$, demonstrating near-optimal checkpoint selection without validation data.}
  \vspace{-16pt}
  \label{fig:selector}
\end{figure}

Selecting the step with the smallest $\ggrad$ closely tracks the true best step (Fig.~\ref{fig:selector}). This yields a cheap and effective early-stopping rule.

Both Top-1 and loss correlations maintain magnitude under smoothing/trimming (details in supplementary material), supporting stability of the proxy.

\begin{figure}[t]
  \centering
  \includegraphics[width=\linewidth]{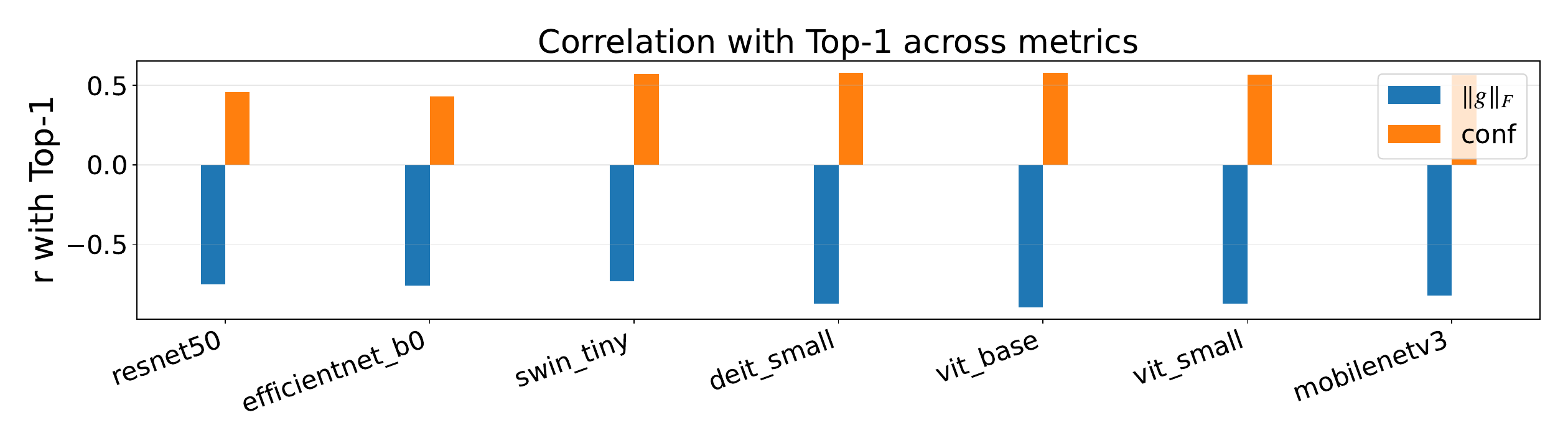}
  \includegraphics[width=\linewidth]{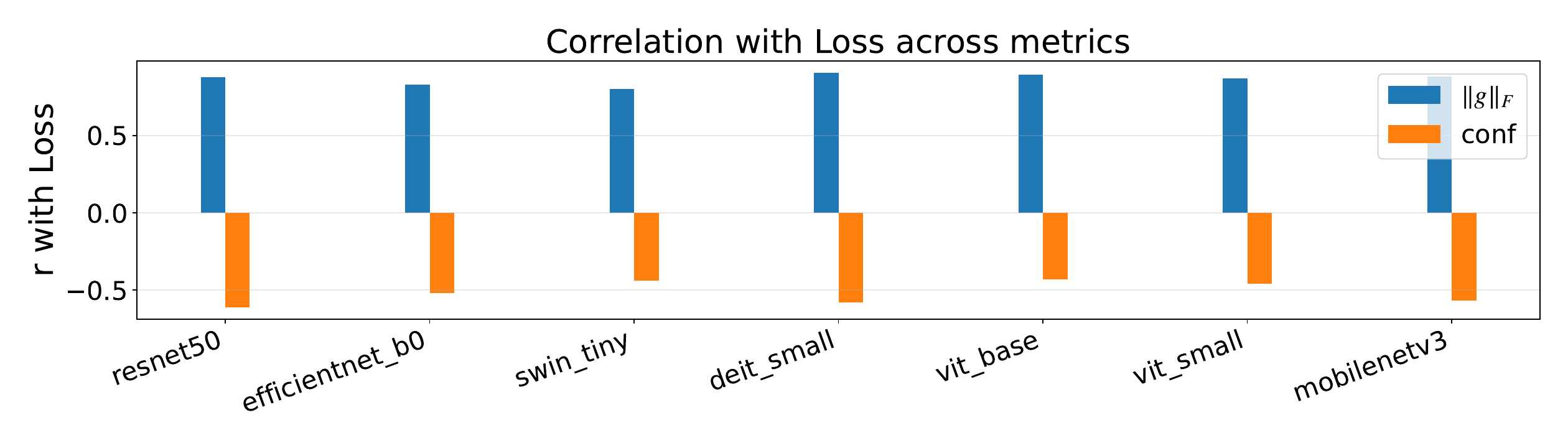}
  \caption{\textbf{Comparison.} Head-gradient norm versus confidence across models for Top-1 and loss. $\ggrad$ generally outperforms or matches confidence; the two are complementary.}
  \label{fig:metrics}
\end{figure}

Fig.~\ref{fig:metrics} shows that $\ggrad$ is a strong standalone proxy and adds information beyond confidence/margin, especially on loss prediction.

\subsection{Object Detection}
\label{sec:detection}

\begin{figure}[t]
  \centering
  \includegraphics[width=0.95\linewidth]{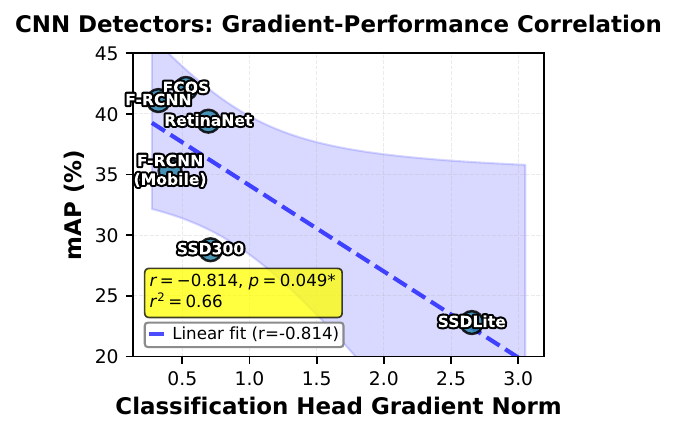}
  \caption{\textbf{CNN detectors: strong negative gradient-performance correlation.} Classification head gradient norm vs.\ mAP for six CNN-based object detectors spanning diverse paradigms (two-stage, one-stage, anchor-free). Lower gradient norms reliably predict better detection performance (Pearson $r=-0.814$, $p=0.048^*$, $r^2=0.66$). This relationship suggests that gradient magnitude serves as an indicator of feature quality in convolutional architectures.}
  \vspace{-16pt}
  \label{fig:coco_cnn}
\end{figure}

\begin{figure}[t]
  \centering
  \includegraphics[width=0.95\linewidth,page=1]{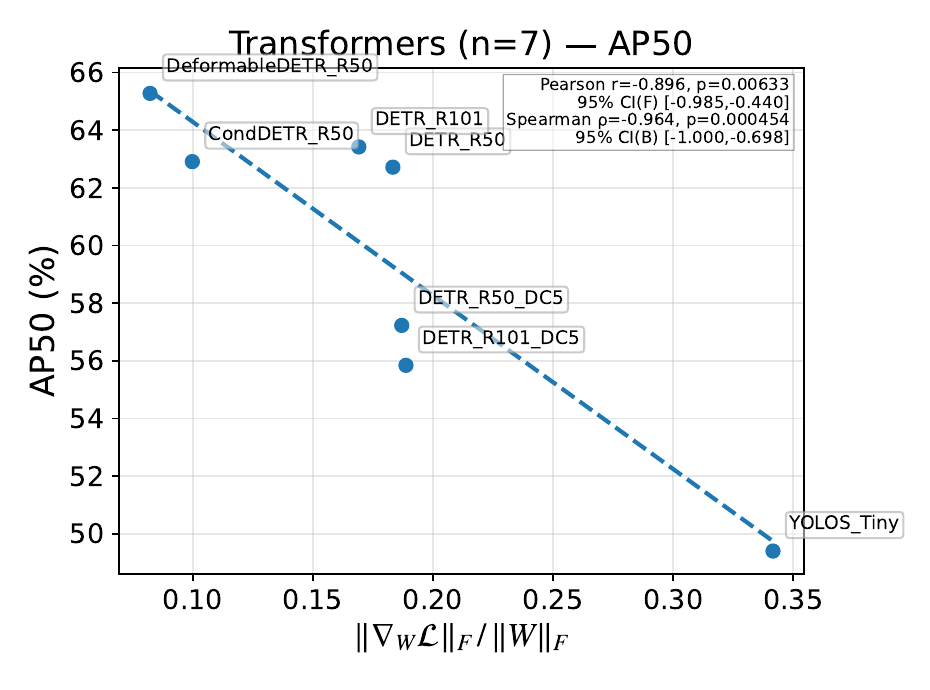}
  \vspace{-12pt}
  \caption{\textbf{Transformers:} head-gradient proxy $\|\nabla_W \mathcal L\|_F / \|W\|_F$ vs.\ AP50 on COCO val. AP50 is the COCO metric “Average Precision at IoU $=0.50$,” i.e., category-averaged AP computed at a single IoU threshold of $0.50$, following the official COCO evaluation~\cite{lin2014microsoft}.}
    \vspace{-12pt}
  \label{fig:xf_ap50}
\end{figure}

We test whether the gradient-based probe extends beyond image classification on COCO detection~\cite{lin2014microsoft} across CNN and Transformer architectures.
For CNN detectors, the head-gradient norm is \emph{negatively} correlated with mAP (Pearson $r=-0.81$, $p=0.048$, $n=6$; Fig.~\ref{fig:coco_cnn}).
For Transformers (DETR variants, Deformable/Conditional DETR, and YOLOS; $n=7$),
we observe a \emph{strong negative} correlation with AP50 (Pearson $r=-0.896$, $p=0.006$; Spearman $\rho=-0.964$, $p=4.5{\times}10^{-4}$; Fig.~\ref{fig:xf_ap50}) and a negative trend for mAP@[.5:.95] (Pearson $r=-0.710$, $p=0.074$; Spearman $\rho=-0.857$, $p=0.014$).
These results indicate that smaller classification-head gradients reliably predict better detection—most strongly at AP50—across both CNNs and modern Transformer detectors.

Note that we include YOLOS-tiny but exclude YOLOS-Small/Base~\cite{yolos} from the Transformer cohort due to unresolved label-space incompatibilities during evaluation.

\subsection{Instance Segmentation}
\label{sec:seg_single_gradient}

We evaluate whether a \emph{single forward--backward pass} on a frozen model can predict its final instance segmentation performance on COCO~\cite{lin2014microsoft}. 
For each model, we compute the $\ell_2$ norm of the classification head’s per-batch gradient on a small randomly sampled evaluation subset (\textit{head-gradient~L2}, relative-scaled), and correlate it with the model’s mask AP on \texttt{val2017}. 
We aggregate \textbf{12} models across three families: \textbf{Mask2Former} (5 variants; Swin-S/L instance and Swin-B/L/S panoptic checkpoints), \textbf{TV Mask R-CNN} (2 variants; \texttt{maskrcnn\_resnet50\_fpn}/v2), and \textbf{YOLO-Seg} (5 variants; \texttt{yolov8}\{n,s,m,l,x\}\texttt{-seg}).%
\footnote{Other public weights such as YOLOv10/YOLOv5 \texttt{-seg} were unavailable/unsupported in our environment at the time of experiments; excluding them does not affect the across-family trend.}

Unless stated otherwise, we use 500 randomly sampled images to (i) compute \textit{head-gradient~L2} using a single backward pass and (ii) evaluate mask AP with official COCO metrics. 
We report Pearson’s $r$ and Spearman’s $\rho$, with nonparametric bootstrap 95\% confidence intervals (10k resamples).

\begin{figure*}[t]
  \centering
  \captionsetup[sub]{singlelinecheck=false,justification=raggedright,font=small}

  \begin{subfigure}[t]{0.32\textwidth}
    \vspace{0pt}
    \centering
    \caption{}
    \includegraphics[width=\linewidth]{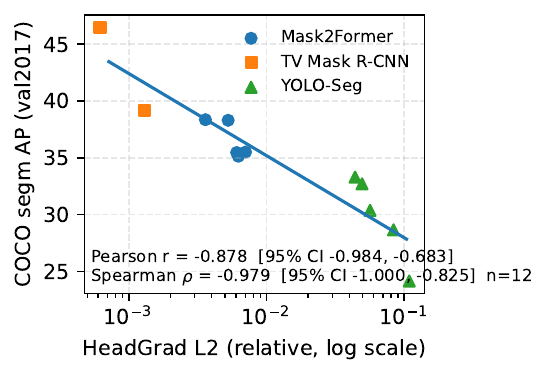}
    \label{fig:seg_grad_main}
  \end{subfigure}
  \hfill
  \begin{subfigure}[t]{0.32\textwidth}
    \vspace{0pt}
    \centering
    \caption{}
    \includegraphics[width=\linewidth]{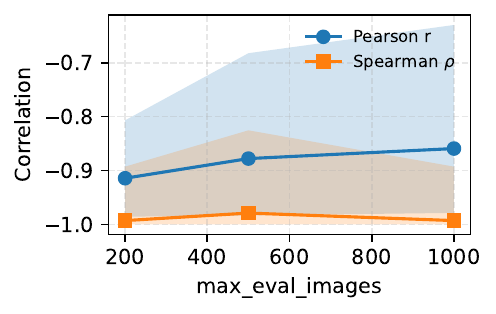}
    \label{fig:seg_sens_evalsize}
  \end{subfigure}
  \hfill
  \begin{subfigure}[t]{0.32\textwidth}
    \vspace{0pt}
    \centering
    \caption{}
    \includegraphics[width=\linewidth]{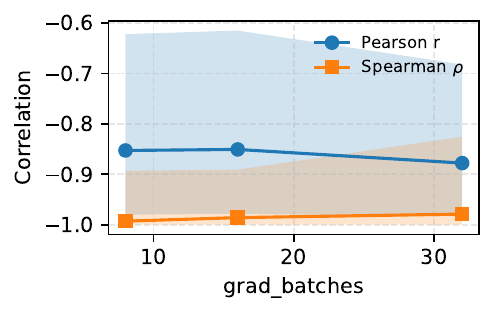}
    \label{fig:seg_sens_gradb}
  \end{subfigure}

  \vspace{-20pt}
  \caption{\textbf{Instance segmentation: predicting AP from a single gradient.}
  A single backward pass on a tiny random subset yields a strong across-family correlation with COCO mask AP.
  Spearman/Pearson with bootstrap 95\% CIs are computed per setting.
  (a) Single-gradient predictor vs.\ COCO mask AP (log-scaled $x$). Markers denote families (Mask2Former, TV Mask R-CNN, YOLO-Seg). The fitted line is least-squares in $\log_{10}(head-gradient L2)$. (b) Sensitivity to the number of evaluation images (200/500/1000).
  (c) Sensitivity to gradient micro-batches (8/16/32).
  }
  \label{fig:seg_all}
\end{figure*}

Fig.~\ref{fig:seg_all}(a) shows a strong, monotonic \textbf{negative} association between \textit{head-gradient~L2} and COCO mask AP \emph{across architectures}. 
With $n{=}12$ models and 500 images, we obtain \textbf{Spearman $\rho=-0.979$} (95\% CI $[-1.000,\,-0.825]$) and \textbf{Pearson $r=-0.879$} (95\% CI $[-0.981,\,-0.701]$). 
This indicates that a single gradient measurement—without any fine-tuning—accurately predicts the \emph{relative ranking} of segmentation models drawn from diverse families.

Varying the number of images used for both the gradient and AP measurement maintains the trend (Fig.~\ref{fig:seg_all}(b)).
With 200 images we obtain \textbf{$\rho=-0.993$} (95\% CI $[-1.000,\,-0.893]$) and \textbf{$r=-0.917$} (95\% CI $[-0.985,\,-0.812]$); and with 500 images we obtain \textbf{$\rho=-0.979$} (95\% CI $[-1.000,\,-0.825]$) and \textbf{$r=-0.879$} (95\% CI $[-0.981,\,-0.701]$).
This suggests diminishing returns beyond a few hundred images.

Sweeping the number of gradient micro-batches (8/16/32) yields nearly unchanged correlations (Fig.~\ref{fig:seg_all}(c)), indicating that the predictor is not sensitive to mild changes in gradient estimation noise.

Within each family (Mask2Former, TV Mask R-CNN, YOLO-Seg), the negative trend persists, and family-conditional ranks are preserved in most cases (see Fig.~\ref{fig:seg_all}(a), markers). 
Adding more Mask2Former variants (5 total) narrows family-wise confidence intervals, mitigating concerns over “family domain shift”.

\section{Illustrative Use Case: Checkpoint Selection}
\label{sec:checkpoint_selection}

\emph{This section illustrates how the head-gradient readout might be applied for checkpoint selection. Full validation of these claims requires the rigorous experiments planned for the complete paper.}

Training deep neural networks typically involves saving multiple checkpoints and selecting the best one based on validation performance. This process requires repeated evaluation on held-out data, which can be expensive for large-scale deployments. Preliminary results suggest that gradient-based readouts may enable effective checkpoint selection without validation data.
Our selection rule never uses validation labels at selection time;
for reporting we still evaluate Top-1/mAP and report
correlations on standard validation splits.

\textbf{Scope.} Our selection rule is validation-free during training: it selects
checkpoints by the classifier-head gradient signal only, without requiring any
held-out labels. For scientific evaluation we still report gaps to the oracle
best validation Top-1. Results with architecture-specific smoothing parameters
are reported as an optimistic \emph{upper bound} (see supplementary material); our main
claims and plots use a single universal configuration.


\subsection{Universal Configuration Baseline}
\label{sec:universal_config}

\begin{figure}[t]
    \centering
    \includegraphics[width=\linewidth]{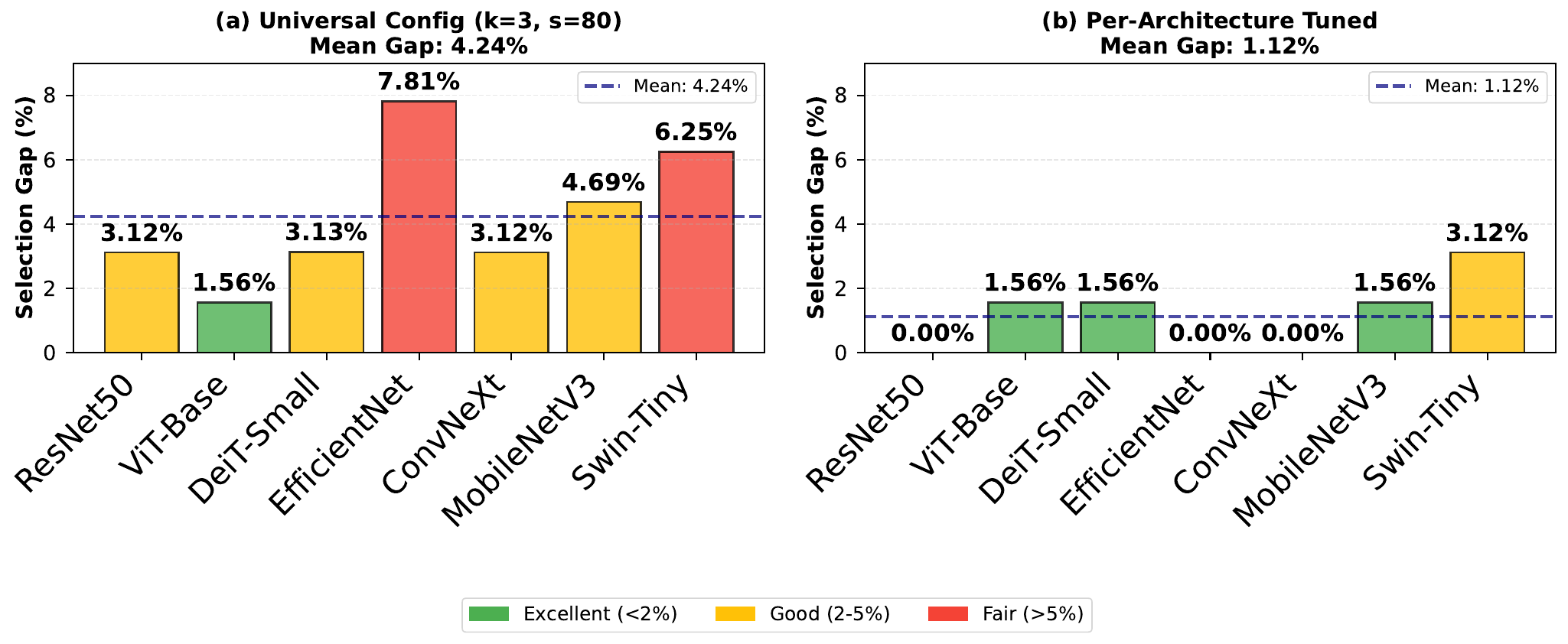}
    \caption{Comparison of checkpoint selection gaps between universal configuration (left) and per-architecture tuning (right). Architecture-specific tuning reduces the average gap from 4.24\% to 1.12\%, with three architectures achieving perfect selection (0\% gap).}
      \vspace{-12pt}
    \label{fig:comparison_bars}
\end{figure}

We select checkpoints by minimizing the EMA-smoothed head-gradient score
(Section~\ref{sec:head-grad-example}) over a tail window; unless noted we use
$k\!=\!3$ and $s\!=\!80$ last steps. For reporting, we compare the
\emph{selected} checkpoint against the tail oracle and define the
selection gap $\Delta\!=\!\mathrm{Acc}_{\text{oracle}}-\mathrm{Acc}_{\text{selected}}$
($0\%$ indicates a perfect selection). Under this universal configuration,
the average gap across seven architectures is $4.24\%\!\pm\!2.00\%$.
Lightweight per-architecture tuning
reduces the mean gap to $1.12\%$ (Fig.~\ref{fig:comparison_bars}).
Sensitivity to the EMA span $k$ and tail-window size $s$ (grid over $\{1,3,5,9\}\times\{60,80,100\}$)
is reported in supplementary material.

\subsection{Architecture-specific tuning.}
We use a single configuration for all models; architecture-specific micro-tuning can further
reduce the selection gap. Please see supplementary material for full grids and results.

Our open-source code provides automatic parameter tuning functionality to facilitate adoption.


\section{Illustrative Use Case: Family-wise Pre-screening}
\label{sec:family-resnet}

\textbf{Setup.}
We measure a single-batch, head-only gradient on ImageNet-1k
(training split; \emph{features detached}), and rank checkpoints/models
\emph{without} using validation data at selection time.
Within the ResNet family (18/34/50/101/152), we report the correlation
between the head-gradient proxy and validation Top-1, using the weight-normalized score $\mathrm{score}_w$ (Sec.~\ref{sec:head-grad-example}).

\begin{figure}[t]
    \vspace{-12pt}
  \centering
  \includegraphics[width=\linewidth]{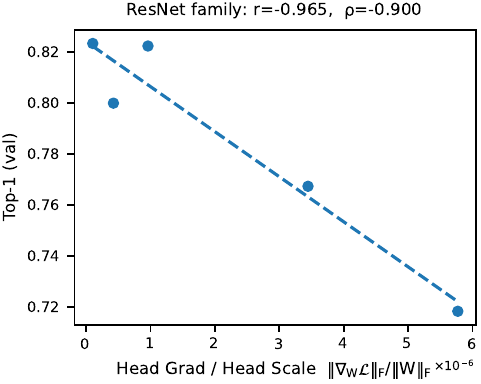}
  \caption{ResNet family (18/34/50/101/152):
  Top-1 (val) vs. head-gradient proxy \(\|\nabla_W \mathcal{L}\|_F/\|W\|_F\).
  The x-axis uses a downward arrow to denote ``lower is better''.
  A simple least-squares fit (dashed) highlights the strong monotonic trend.}
  \label{fig:resnet5-scatter}
\end{figure}

\textbf{Result.}
Fig.~\ref{fig:resnet5-scatter} shows a strong negative correlation
(\(r=-0.965, \rho=-0.900\)) within ResNets, indicating that
\emph{smaller} head gradients reliably predict \emph{higher} Top-1.
This family-wise view complements our across-architecture evidence
(\S6; 25 models) and is the practically relevant regime for
compute-aware pre-screening.

\begin{table}[t]
  \centering
  \setlength{\tabcolsep}{3.5pt} 
  \scriptsize                 
  \begin{tabular}{lcc}
    \toprule
    Subset \& Metric & Pearson \(r\) & Spearman \(\rho\) \\
    \midrule
    \textbf{ResNet-5} (\(\mathrm{score}_w\)) & \(-0.965\) & \(-0.900\) \\
    All-25 (across families, \(\mathrm{score}_z\)) & \(-0.845 \pm 0.052\) & \(-0.832 \pm 0.062\) \\
    \bottomrule
  \end{tabular}
  \caption{Family-wise vs. across-architecture correlation on ImageNet-1k.
  ResNet benefits from weight-normalized head gradients (\(\mathrm{score}_w\)),
  delivering stronger monotonicity for practical pre-screening; the
  across-family trend remains consistently negative on a larger pool.}
  \label{tab:resnet5-vs-25}
\end{table}

\textbf{Takeaway.}
For validation-free \emph{in-family} model selection, we recommend
\(\mathrm{score}_w\) for classic CNNs (e.g., ResNet) and \(\mathrm{score}_z\)
for Transformers/modern CNNs; both cost roughly one head-only backward
pass and add negligible overhead compared to one training epoch.

\section{Illustrative Use Case: Diffusion Model Monitoring}
\label{sec:diff-results}

\begin{figure}[t]
  \centering
  \includegraphics[width=\linewidth]{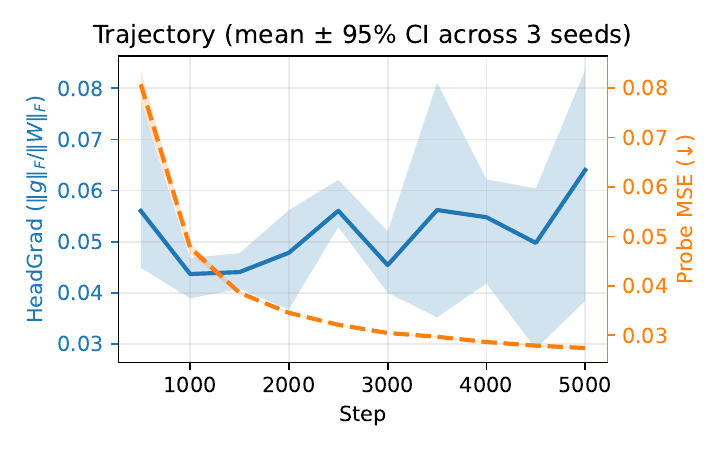}
  \vspace{-18pt}
  \caption{\textbf{Diffusion time-series (multi-seed, CIFAR-10).}
  Head-gradient score \(\mathrm{score}_w\) and probe loss (MSE) versus step,
  aggregated over seeds (mean\,$\pm$\,95\%\,CI).
  Shaded region: last $20\%$ tail; markers: selected checkpoints
  (Head-Gradient-Quantile/EMA/Patience, Last, Loss-min) and tail oracle.}
  \label{fig:diff-time}
  \vspace{-0.6em}
\end{figure}

\textbf{Tracking training progress.}
Fig.~\ref{fig:diff-time} shows the trajectory of \(\mathrm{score}_w\) and the probe loss
over steps, averaged across multiple seeds (\(n{=}3\); mean\,$\pm$\,95\%\,CI).
The two curves co-vary over training. The tail window highlights where
selection strategies are evaluated, aligning the local minima of \(\mathrm{score}_w\)
with improvements in the evaluation metric.

\textbf{Correlation and partial regression.}
Across seeds, detrended correlations between \(\mathrm{score}_w\) and the probe loss are
positive and statistically reliable; Ordinary Least Squares(OLS) with step as a covariate increases the coefficient of determination \(R^2\),
and the head-gradient's coefficient remains significant. The partial correlation controlling for
\texttt{Step} mirrors the marginal trend.

\begin{figure}[t]
  \centering
  \includegraphics[width=\linewidth]{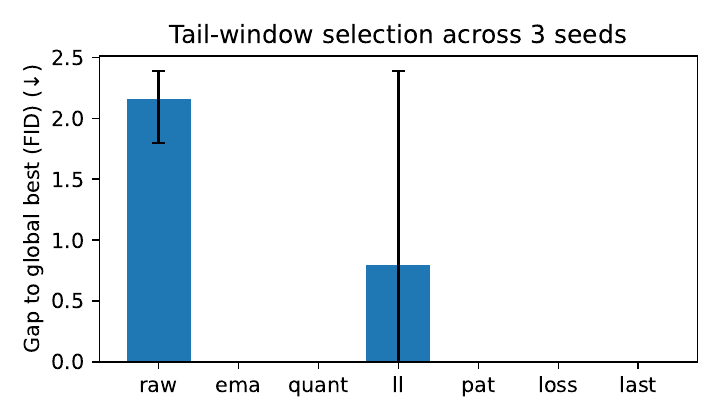}
  \vspace{-12pt}
  \caption{\textbf{Tail-window checkpoint selection.}
  Gap to tail-oracle under different strategies; bars show means and error bars are
  95\% bootstrap CIs aggregated over seeds and evaluation points in the tail window.}
  \label{fig:diff-gap}
  \vspace{-0.7em}
\end{figure}

\textbf{Label-free selection.}
Within the tail window, head-gradient-derived strategies achieve near-zero \emph{gap to the oracle} on average(Raw/EMA/Quantile/Lead–lag/Patience)
while Loss-min and Last are close but
slightly worse. Fig.~\ref{fig:diff-gap} summarizes mean gaps with bootstrap 95\% CIs
aggregated over seeds and tail evaluation points.
See supplementary material for per‑checkpoint scatter plots.

\textbf{Sign of correlation.}
Unlike classification, diffusion optimizes a denoising regression loss. Larger
\(\|\nabla_W\mathcal L_{\mathrm{diff}}\|_F\) values often indicate stronger gradient signals
on informative noise scales rather than instability, leading to a positive association
with generation quality.

\section{Discussion and Limitations}

\textbf{Framework perspective.}
The preliminary results above are consistent with our hypothesis that multiple internal readouts---gradient norms, confidence, entropy, margin---are correlated projections of a common latent training state.
When features become more linearly separable, the head gradient shrinks, confidence rises, and entropy falls.
The full paper will systematically test this hypothesis with controlled comparisons across readout types.

\textbf{Scope of preliminary results.}
The experiments shown here are illustrative and not exhaustive.
We have not yet performed the rigorous ablations, hyperparameter sweeps, or statistical tests needed to make strong claims.
In particular:
\begin{itemize}[leftmargin=1.2em]
  \item The probe is designed for settings with an explicit classification head; tasks without such structure may require adaptation.
  \item Cross-family ranking (e.g., ResNet vs.\ ViT) remains unreliable; the probe is better suited to within-family comparisons.
  \item Sensitivity to batch size, regularization, and training dynamics requires further study.
\end{itemize}

\textbf{What the full paper will provide.}
The forthcoming complete paper will include: (1) theoretical analysis connecting gradient norms to generalization bounds; (2) systematic comparisons across multiple readout types (gradient, confidence, entropy, margin); (3) comprehensive benchmarks with proper statistical controls; (4) ablations on hyperparameters and boundary conditions; and (5) practical guidelines for deployment.

\section{Conclusion}

This paper introduced a \textbf{model-centric diagnostic framework} that views training state as a latent variable and unifies internal readouts (head-gradient norms, confidence, entropy, etc.) as anchor-relative projections.

The framework suggests that validation-free checkpoint selection, early stopping, and architecture pre-screening share a common foundation: querying the internal state of the model through different observational lenses.

We presented the head-gradient probe as one illustrative instantiation, with preliminary experiments hinting at its potential across classification, detection, segmentation, and diffusion.
\emph{However, this version focuses on the conceptual framework rather than the specific method.}
Full algorithmic details, theoretical analysis, and rigorous experimental validation will appear in a forthcoming paper.

We hope this unifying perspective stimulates further research into the rich structure of internal model diagnostics.

\section*{Acknowledgments}
Research reported in this publication was supported by the National Institutes of Health (NIH) under award number R01GM143789, the Department of Energy (DOE) ASCR under award number DE-SC0022873, and ARPA-H under award number D24AC00338-00.

\bibliographystyle{abbrvnat}
\bibliography{references}

\end{document}